%% file: main.tex
\documentclass[10pt,twocolumn,letterpaper]{article}
\usepackage[table,xcdraw]{xcolor}
\usepackage{pifont}
\usepackage{arydshln}
\usepackage[normalem]{ulem}
\usepackage{cvpr}    
\usepackage{multirow}
\usepackage[accsupp]{axessibility} 
\input{preamble}
\definecolor{cvprblue}{rgb}{0.21,0.49,0.74}
\usepackage[pagebackref,breaklinks,colorlinks,allcolors=cvprblue]{hyperref}

\def\paperID{5091} 
\def\confName{CVPR}
\def\confYear{2026}

\title{MeanFuser: Fast One-Step Multi-Modal Trajectory Generation and Adaptive Reconstruction via MeanFlow for End-to-End Autonomous Driving}

\author{
Junli Wang$^{1,2,3 }$,
Yinan Zheng$^{3,4}$,
Xueyi Liu$^{1,2}$,  
Zebin Xing$^{1,2}$, 
Pengfei Li$^{4}$,
Kun Ma$^3$,\\
Hangjun Ye$^3$,
Guang Chen$^{3}$, 
Guang Li$^3$,
Long Chen$^{3}$,
Zhongpu Xia$^{1,2}$,
Qichao Zhang$^{1,2 \dagger}$ \\
[2mm]
$^1$~SKL-MAIS, Institute of Automation, Chinese Academy of Sciences \\
$^2$~School of Artificial Intelligence, University of Chinese Academy of Sciences \\
$^3$~Xiaomi EV \quad
$^4$~Institute for AI Industry Research (AIR), Tsinghua University
}

\begin{document}
\maketitle
{
\renewcommand{\thefootnote}{}
\footnotetext{%
  \hspace{-1.8em}%
  \textdagger\ Corresponding author.
  This work is supported by the Beijing Natural Science Foundation-Xiaomi Innovation Joint Fund L253007, and Beijing Natural Science Foundation under Grant 4242052.
  Our code are available at https://github.com/wjl2244/MeanFuser.}
}
\input{sec/0_abstract}
\input{sec/1_intro}
\input{sec/2_method}
\input{sec/3_experiments}
{
    \small
    \bibliographystyle{ieeenat_fullname}
    \bibliography{main}
}

\input{sec/X_suppl}

\end{document}

%% file: sec/0_abstract.tex
\begin{abstract}

Generative models have shown great potential in trajectory planning. Recent studies demonstrate that anchor-guided generative models are effective in modeling the uncertainty of driving behaviors and improving overall performance. However, these methods rely on discrete anchor vocabularies that must sufficiently cover the trajectory distribution during testing to ensure robustness, inducing an inherent trade-off between vocabulary size and model performance. To overcome this limitation, we propose \textbf{MeanFuser}, an end-to-end autonomous driving method that enhances both efficiency and robustness through three key designs. (1) We introduce Gaussian Mixture Noise (GMN) to guide generative sampling, enabling a continuous representation of the trajectory space and eliminating the dependency on discrete anchor vocabularies. (2) We adapt ``MeanFlow Identity" to  end-to-end planning, which models the mean velocity field between GMN and trajectory distribution instead of the instantaneous velocity field used in vanilla flow matching methods, effectively eliminating numerical errors from ODE solvers and significantly accelerating inference. (3) We design a lightweight Adaptive Reconstruction Module (ARM) that enables the model to implicitly select from all sampled proposals or reconstruct a new trajectory when none is satisfactory via attention weights.Experiments on the NAVSIM closed-loop benchmark demonstrate that MeanFuser achieves outstanding performance
without the supervision of the PDM Score and exceptional inference efficiency, offering a robust and efficient solution for end-to-end autonomous driving.

\end{abstract}

%% file: sec/1_intro.tex
\section{Introduction}
\label{sec:intro}

    \begin{figure}[tbp]
        \centering
            \centering
            \begin{subfigure}[b]{0.5\textwidth}
                \centering
                \includegraphics[width=0.97\textwidth]{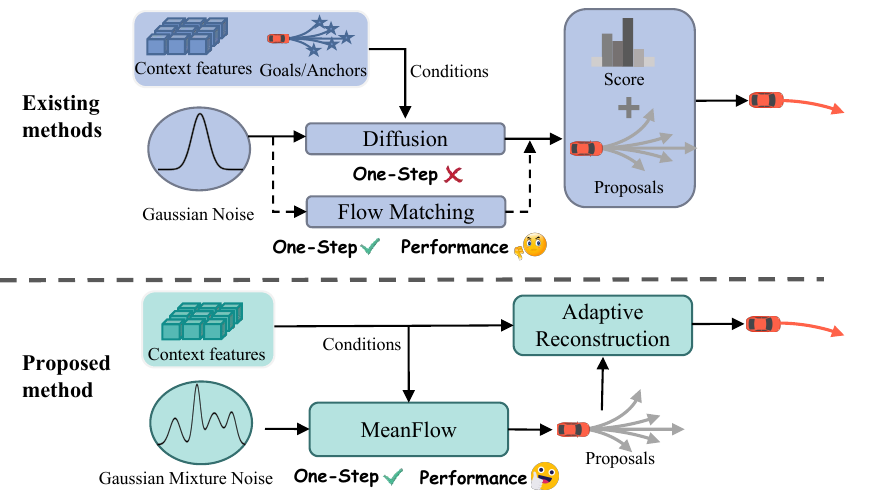}
                \caption{}
                \label{fig:teaser}
            \end{subfigure}
    
            \vspace{0cm} 
    
            \begin{subfigure}[b]{0.45\textwidth}
                \centering
                \includegraphics[width=0.95\textwidth]{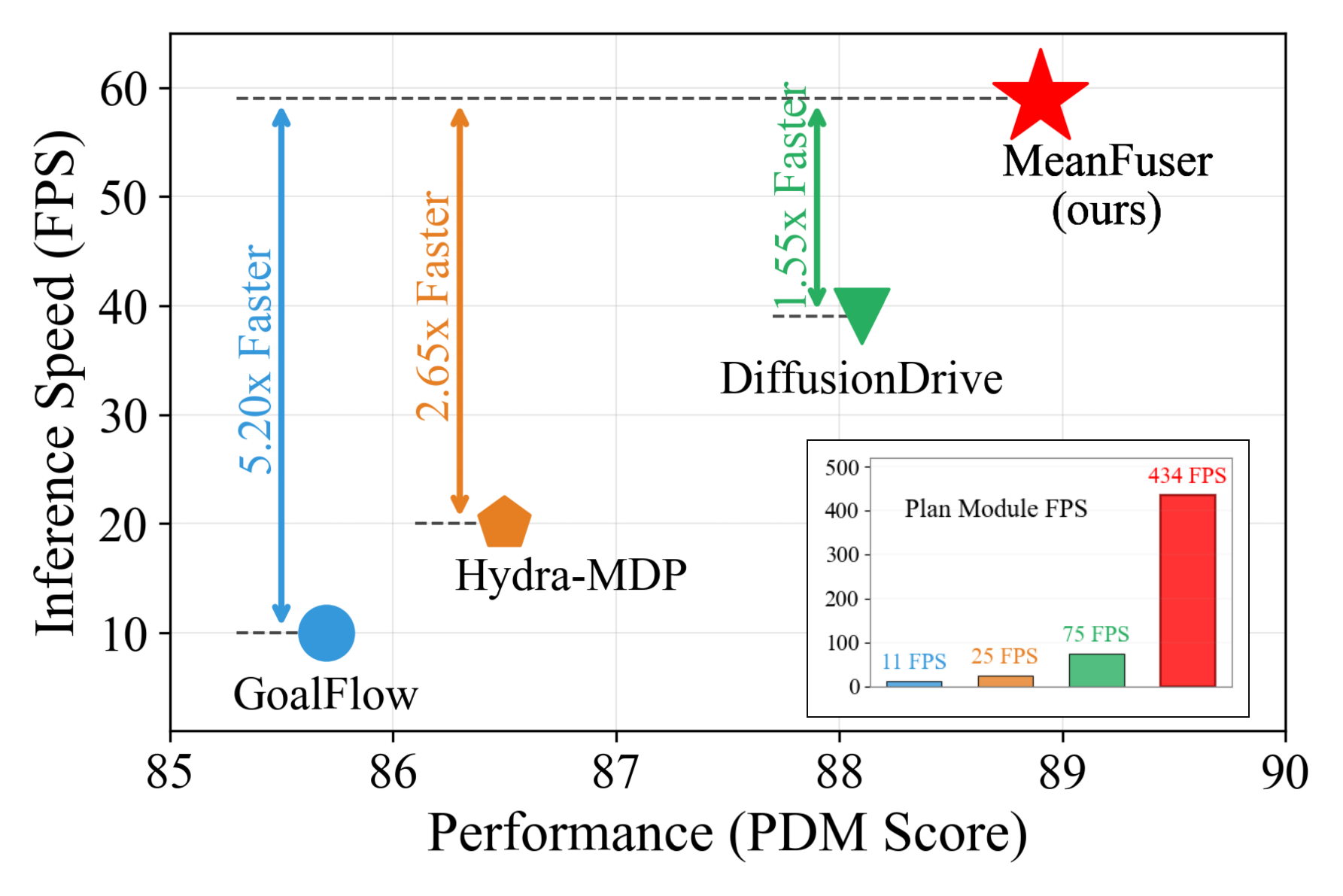}
                \caption{}
                \label{fig:fps}
            \end{subfigure}
        \caption{(a) illustrates the differences between our proposed method and existing generative approaches, highlighting the introduction of Gaussian mixture noise to replace anchor vocabularies, one-step sampling, and the adaptive reconstruction module. (b) shows the advantages of MeanFuser over GoalFlow\cite{goalflow}, Hydra-MDP\cite{Hydra-MDP}, and DiffusionDrive\cite{DiffusionDrive} in terms of closed-loop performance, inference speed and plan module inference speed.}
    \vspace{-0.1in}
    \end{figure}

    \begin{figure*}[ht]
        \begin{center}
        \centerline{\includegraphics[width=1.9\columnwidth]{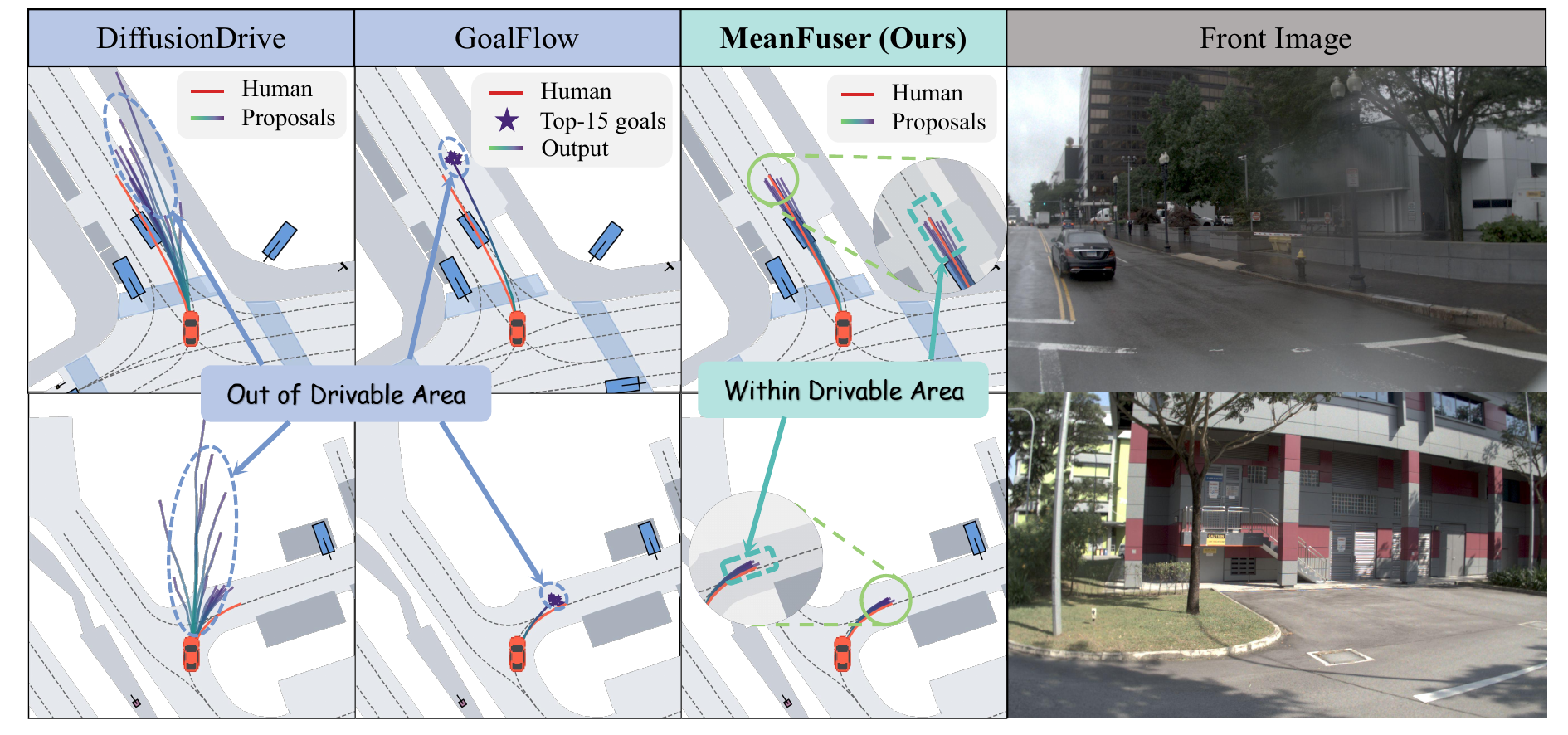}}
        \caption{{Failure scene visualization. Anchor-guided models (GoalFlow, DiffusionDrive) fail due to the inability of discrete vocabularies to cover the entire trajectory space, while our model generates proposals that encompass the optimal trajectories.}}
        \label{fig:scenes_1}
        \end{center}
    \vskip -0.2in
    \end{figure*}

    End-to-end autonomous driving\cite{e2e,transfuser,ST-P3,reasonplan,dual-aeb} has attracted considerable attention as it directly learns the mapping from raw sensor inputs to planned trajectories, thereby avoiding handcrafted rules in rule-based systems and the cumulative errors inherent in modular pipelines\cite{zheng2026planagent}. 

    High-performance frameworks such as TransFuser\cite{transfuser}, UniAD\cite{UniAD}, and VAD\cite{VAD} have demonstrated impressive driving capabilities by learning unimodal trajectories. However, they fail to effectively capture the multi-modal nature and uncertainty of driving behaviors. Methods like VADv2\cite{VADv2} and Hydra-MDP\cite{Hydra-MDP} introduce a trajectory vocabulary and predict the probability distribution of each trajectory under specific scenarios. However, a fixed vocabulary size imposes a trade-off between efficiency and robustness. Large vocabularies slow down inference, while small ones tend to fail under distribution shifts.

    Recently, generative models such as diffusion models\cite{ddpm,ddim,scoremodel} and flow-based models\cite{flowmatching,rectifiedflow,meanflow} have demonstrated remarkable capabilities in capturing complex data distributions, leading to significant breakthroughs in fields like image generation and robot learning\cite{diffusionpolicy,flowpolicy,mp1}. Motivated by their success, researchers have begun exploring the use of diffusion and flow matching frameworks in end-to-end autonomous driving. However, vanilla diffusion models often suffer from mode collapse\cite{cfg++}, resulting in limited trajectory diversity. To address this issue, various approaches have been proposed to improve trajectory diversity. For example, DiffusionDrive\cite{DiffusionDrive} integrates a clustered trajectory vocabulary within an iterative refinement framework; Similarly, GoalFlow\cite{goalflow} predicts the top-k goal points, aggregates them and guides the flow matching process, however, it requires five sampling steps to achieve optimal performance.
    Despite these advancements, existing diffusion and flow matching approaches still suffer from significant limitations. One issue is the need for multiple sampling steps, which trade off inference speed and model performance. Another is their discretizations of the trajectory space, where the discretizations inherently constrain the exploration of the broader trajectory solution space and compromise their performance on test scenarios that fall outside the pre-defined anchor distribution, as shown in Fig.~\ref{fig:scenes_1}. This raises a key question: how can we effectively model the multi-modal nature of driving behaviors while maintaining strong performance \textbf{without relying on a fixed, discrete anchor vocabulary?}

    To tackle the above challenges, we propose \textbf{MeanFuser}, as illustrated in Fig.~\ref{fig:teaser}. First, to address the mode collapse caused by standard Gaussian noise sampling and to eliminate the dependence on a fixed discrete vocabulary, we model the noise distribution using a \textbf{Gaussian Mixture Noise (GMN)}. Unlike vocabulary-based approaches that discretize the trajectory space, GMN provides a continuous representation that covers a broader range of trajectories. Each Gaussian component captures a distinct driving pattern, effectively enhancing the model’s ability to represent multi-modal behaviors. Interestingly, we find that the GMN also brings an additional benefit, different components naturally correspond to diverse driving styles, opening possibilities for customized autonomous driving.

    Second, to accelerate sampling and eliminate the numerical errors introduced by Ordinary Differential Equation (ODE) solvers during the sampling process, we adapt ``MeanFlow Identity"\cite{meanflow} to  end-to-end planning, which directly learns the mean velocity field between the noise distribution and trajectory distribution, replacing the instantaneous velocity field used in vanilla flow matching models.

    Finally, we consider a practical issue: what if all sampled proposals are suboptimal in certain cases? Our solution is to reconstruct high-quality trajectories when necessary. To achieve this, we design an \textbf{Adaptive Reconstruction Module (ARM)} that enables the model to implicitly decide via attention weights, after evaluating all proposals, whether to select an existing trajectory or to regenerate a new one. Unlike methods such as Hydra-MDP\cite{Hydra-MDP} and WoTE\cite{wote}, which score candidate trajectories using evaluation submetrics from benchmark, ARM enhances performance while relying solely on expert trajectories.

    We evaluate MeanFuser on a closed-loop benchmark NAVSIM. As shown in Fig.~\ref{fig:fps}, Under the same perception backbone, MeanFuser achieves \textbf{89.0 PDMS} on NAVSIMv1 and \textbf{89.5 EPDMS} on NAVSIMv2, surpassing previous methods. Moreover, owing to its one-step sampling design, MeanFuser achieves \textbf{59 frames per second (FPS)}, with inference speeds 5.20×, 2.65×, and 1.55× faster than GoalFlow, Hydra-MDP, and DiffusionDrive, respectively. Our main contributions are summarized as follows:

    \begin{itemize}
        \item We propose \textbf{MeanFuser}, the first framework that introduces the MeanFlow paradigm into end-to-end autonomous driving, achieving an  effective balance between one-step sampling and planning performances.

        \item We propose Gaussian mixture noise \textbf{(GMN)} modeling to capture diverse driving modes without relying on fixed anchor vocabulary.

        \item We design an adaptive reconstruction module \textbf{(ARM)} to handle suboptimal proposals, enhancing robustness and trajectory quality in complex scenarios.

        \item We demonstrate that MeanFuser outperforms existing imitation learning based approaches on NAVSIM closed-loop Benchmark, achieving superior results with a rule-free and image-only design.
    \end{itemize}


\section{Related Work}

    \subsection{End-to-End Autonomous Driving}
        End-to-end autonomous driving systems take sensory inputs and ego-vehicle states as inputs and aim to directly generate driving trajectories through neural optimization\cite{zheng2025world4drive, yang2026worldrft}. Early works such as Transfuser\cite{transfuser} focused on fusing heterogeneous modalities such as camera and LiDAR, while introducing auxiliary tasks to enhance supervision during action-space training. UniAD\cite{UniAD} further structured the entire framework as a planning-oriented transformer pipeline, enabling information flow across multiple specialized modules. Building on this line of research, VAD\cite{VAD} replaced dense bird’s-eye-view representations with lightweight vectorized scene abstractions, which significantly improved efficiency. Its successor, VADv2\cite{VADv2}, reformulated trajectory regression as a discrete trajectory-token selection problem, choosing the optimal trajectory from a learned vocabulary. ParaDrive\cite{paradrive} departed from the traditional serial end-to-end paradigm and adopted a parallel framework, allowing mapping, prediction, and planning to proceed simultaneously. Unlike the aforementioned methods, we use powerful generative models to model the trajectory distribution.

    \subsection{Diffusion and Flow-Based Generative Models for Trajectory Planning}
        
        Driving behaviors are inherently multimodal since in complex traffic scenarios, multiple feasible trajectories may exist. To capture this stochasticity, recent works adopt diffusion\cite{chen2024boosting,li2024generalizing,chen2025conrft} or flow matching\cite{flowmatching,rectifiedflow} generative paradigms, training with noise injection and denoising inference to sample diverse yet plausible trajectories\cite{DiffusionDrive,consistencydrive,fang2025consistency,goalflow}. MotionDiffuser\cite{motiondiffuser} was among the first to introduce diffusion-based trajectory generation in autonomous driving. Later, Diffusion-ES\cite{diffusion-es} refined the generated trajectories through iterative optimization that alternates between generation and evaluation. Diffusion Planner\cite{diffusion-planner} incorporated scene context through carefully designed conditioning mechanisms that effectively guided the diffusion process. Moreover, DiffusionDrive\cite{DiffusionDrive} introduced noise perturbations to clustered trajectory prototypes, effectively encoding prior trajectory distributions. In contrast, GoalFlow\cite{goalflow} employed flow matching for improved stability and utilized goal points to explicitly guide trajectory generation. We improve the model's performance and robustness without relying on anchor-based guidance by introducing Gaussian Mixture Noise sampling and an Adaptive Reconstruction Module, while requiring only one-step sampling.

    \subsection{Candidate Trajectory Proposal Evaluation and Selection }
        
        Due to the complexity of driving environments, end-to-end planners often generate a set of candidate trajectory proposals followed by a selection stage. The selection mechanism can be rule-based or learning-based. For instance, SparseDrive\cite{sparsedrive} performed trajectory rollouts conditioned on environmental predictions and evaluated them using handcrafted scoring functions. Similarly, Diffusion-ES\cite{diffusion-es} applied map-based scoring rules to filter diffusion-generated trajectories. Learning-based selectors such as DiffusionDrive\cite{DiffusionDrive} and WoTE\cite{wote} adopt data-driven scoring mechanisms. DiffusionDrive jointly learned both trajectory generation and action-value estimation within the diffusion framework, while WoTE introduced a learned world model to simulate environment dynamics and assess candidate trajectories. These approaches bridge the gap between stochastic trajectory generation and decision-level optimization. We eliminate direct trajectory evaluation by introducing the Adaptive Reconstruction Module, allowing the model to  implicitly decide whether to select from the candidates or regenerate a better trajectory based on the proposed candidates.


\begin{figure*}[ht]
    \begin{center}
    \centerline{\includegraphics[width=2\columnwidth]{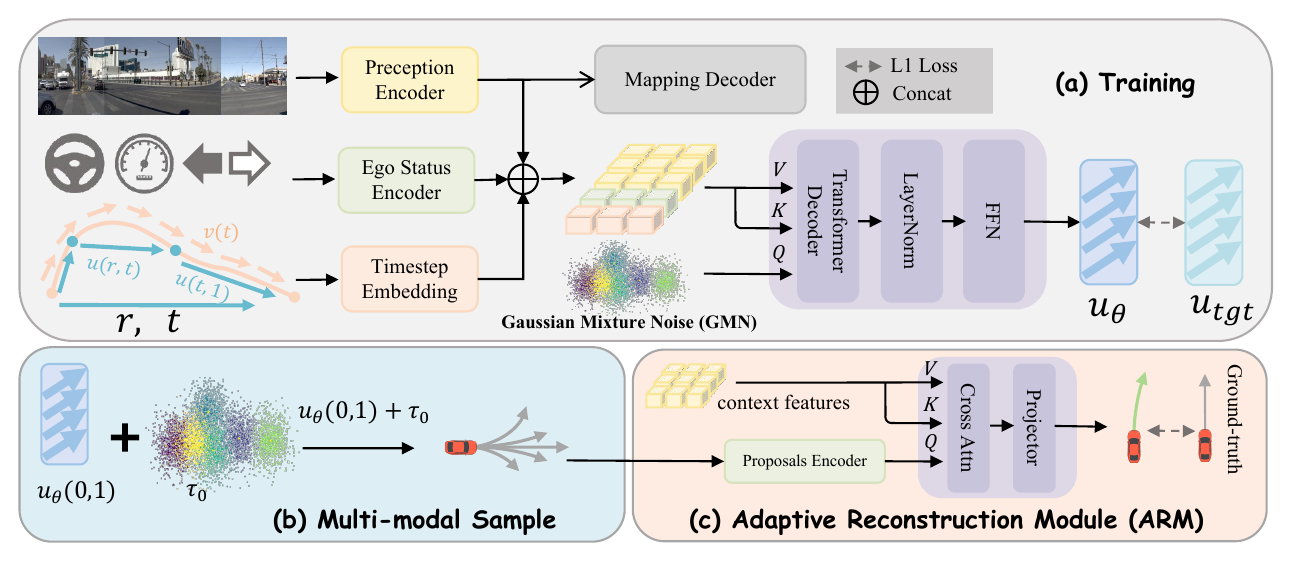}}
    \caption{Overall architecture of MeanFuser. \textbf{Training}: During training, both the images and ego-vehicle states are encoded into context features, with auxiliary supervision from mapping tasks. The model is conditioned on these context features to learn the average velocity field $u_{\theta}$ over the time interval $r$ and $t$. \textbf{Multi-Modal Sample}: Noise samples are drawn from Gaussian Mixture Noise, and the one-step sampling formulation is then applied to generate diverse multi-modal trajectories. \textbf{Adaptive Reconstruction Module}: The sampled multi-modal trajectories are encoded and fused with the context features through cross-attention, after which a Projector outputs the final planning trajectory.}
    \label{fig:framework}
    \end{center}
    \vskip -0.2in
\end{figure*}

\section{Preliminary}

    \subsection{Problem Formulation}

        End-to-end autonomous driving systems take raw sensor data as input and directly predict future driving trajectories. The observational information $\mathcal{O}_i$ at time step $i$ typically comprises multi-view camera images $\mathcal{I}_i \in \mathbb{R}^{H \times W \times C}$, ego-vehicle state information $\mathcal{S}_i$ (including velocity, acceleration), and high-level driving commands $\mathcal{C}_i$. Given the observation inputs $\mathcal{O}_i$, the objective is to predict a future trajectory \(\hat{\tau}=\{\ (x_i,y_i,\theta_i) \}_{i=1}^{T_f}\), where \(x_i,y_i,\theta_i \) represent the position coordinates and heading angle in the ego-vehicle coordinate system at time step $i$, $T_f$ denotes the planning horizon. During training, the model learns from expert demonstration trajectories \(\tau\) to optimize its parameters.

    \subsection{Flow-Based Model}

        We first introduce the foundational concepts of flow matching. Standard flow matching\cite{flowmatching} frameworks aim to construct a continuous probability path $\{{z_t\}}_{t\in[0,1]}$ between a simple prior distribution $p_0$ (e.g., Gaussian noise) and the complex data distribution $p_1$, governed by the boundary conditions $z_{t=0} = p_0$ and $z_{t=1} = p_1$, \(z_t\) is called a flow. This is achieved by learning an instantaneous velocity field $v(z_t, t)$. A sample $x_0 \sim p_0$ is then transformed into a sample $x_1 \sim p_1$ by solving the ordinary differential equation:
        \begin{equation}
            \begin{cases}
            \frac{d z_t}{dt} = v_\theta\!\bigl(z_t, t\bigr) \\
            z_0 = x_0.
            \end{cases}
        \end{equation}
        
        However, a critical limitation persists: even if the marginal probability path $z_t$ is linear, which often constructible as \(z_t = (1-t)x_1 + tx_0\), the learned velocity field $v_\theta\!\bigl(z_t, t\bigr)$ is not guaranteed to produce straight trajectories for individual samples\cite{meanflow}. This inherent curvature forces the use of small step sizes and multiple function evaluations (NFEs) during ODE solvers to minimize discretization error, which significantly hinders sampling efficiency.

        MeanFlow\cite{meanflow} addresses this limitation by abandoning the modeling of instantaneous velocity fields and instead directly modeling the mean velocity field $u(z_t, r, t)$, which is defined as the total displacement between time steps t and r divided by the time interval:
        \begin{equation}
            u(z_t, r, t) \triangleq \frac{1}{t-r} \int_{r}^{t} v(z_\tau, \tau) \, d\tau.
        \end{equation}
        By differentiating and rearranging with respect to \(t\), one can derive the “MeanFlow Identity”:
        \begin{equation}
            u(z_t, r, t) = v(z_t, t) - (t-r)\frac{d}{dt}u(z_t, r, t),
        \end{equation}
        where the differential term can be rewritten as:
        \begin{equation}
            \frac{d}{dt}u(z_t, r, t) = v(z_t, t)\partial_zu + \partial_tu.
        \end{equation}
        This identity establishes a precise relationship between the instantaneous velocity $v(z_t,t)$ and the mean velocity $u(z_t,r,t)$, enabling the development of a novel training objective that explicitly encourages straight sample trajectories. By learning the mean velocity field $u$ directly, the model can generate high-quality samples in a single step, as the straight-line trajectory from $x_0$ to $x_1$ is given by:
        \begin{equation}
            x_1 = x_0 + 1 \cdot u_{\theta}(x_0, 0, 1).
        \end{equation}
            The objective of the model training is to minimize the following loss function:
        \begin{equation}
            \mathcal{L}(\theta) = \mathbb{E} \| u_\theta(z_t, r, t) - \text{sg}(u_{\text{tgt}}) \|_2^2.
        \end{equation}
        The sg(\(\cdot\)) operator refers to the stop-gradient operator, which prevents gradients from being propagated through the target \(u_{\text{tgt}}\) during backpropagation, ensuring that only the model output is updated. The target mean velocity field \(u_{\text{tgt}}\) is defined as:
        \begin{equation}
            u_{\text{tgt}} = v(z_t, t) - (t - r) \left( v(z_t, t) \partial_z u_\theta + \partial_t u_\theta \right),
        \end{equation}
        where the velocity field \( v(z_t, t) \) is given by \( v(z_t, t) = x_0 - x_1 \). During training, we compute the {Jacobian-vector product (JVP)} of the function \( u(z_t, r, t) \) using \texttt{torch.autograd.functional.jvp}. For this, we use the tangent vector \( [v, 0, 1] \), which allows us to obtain the derivatives \( \partial_z u_\theta \) and \( \partial_t u_\theta \).

\section{Method}
     The overall pipeline of MeanFuser is illustrated in Fig.~\ref{fig:framework} of our model architecture, it consists of three main components: the model training, the multi-modal trajectoriers sample, and the adaptive reconstruction module.

    \subsection{Scene Context Encoder}

    The scene encoder consists of two components: an image encoder $\mathcal{E}_I$ and a ego-vehicle state encoder $\mathcal{E}_s$, which are used to extract high-dimensional key features specific to the scene,
    \begin{equation}\label{eqn-1}
        \mathbf{c_{bev}} = \mathcal{E}_I(\mathcal{I}_i), \quad \mathbf{c_s} = \mathcal{E}_s(\mathcal{S}_i, \mathcal{C}_i).
    \end{equation}
    
    During training, an auxiliary decoder is employed to decode information of the lane map from $\mathbf{c_{bev}}$, with the corresponding loss functions $\mathcal{L}_{\text{map}}$ applied to supervise semantic learning and accelerate model convergence. The extracted scene features $\mathbf{c} = \{\mathbf{c_{bev}}, \mathbf{c_s}\}$ serve as scene-specific conditional inputs for multi-modal trajectories sampling.

    \subsection{Gaussian Mixture Noise}

    Unlike diffusion-based models, flow-based models do not require the prior distribution \( p_0 \) to follow standard Gaussian noise. In this work, to eliminate reliance on discrete vocabularies, we set \( p_0 \) as a Gaussian Mixture Noise distribution, guiding the model to generate multi-modal driving trajectories.

    Concretely, we normalize expert demonstrations in the training set. For each trajectory \(\tau_j\), we compute
    \begin{equation}
    \Delta \tau_j = \{(\tau_j)\}_{t=1}^T - \{(\tau_j)\}_{t=0}^{T-1}.
    \end{equation}
    We further compute \(\Delta \tau_{\text{mean}}\), \(\Delta \tau_{\text{max}}\), and \(\Delta \tau_{\text{min}}\) across all trajectories and all time steps:
    \begin{equation}
    \begin{cases}
    \Delta \tau_{\text{mean}} = \underset{j,t}{\text{mean}} ((\Delta \tau_j)_t)\\
    \Delta \tau_{\text{max}} = \underset{j,t}{\max} ((\Delta \tau_j)_t)\\
    \Delta \tau_{\text{min}} = \underset{j,t} {\min} ((\Delta \tau_j)_t).
    \end{cases}
    \end{equation}
    
    Each trajectory is then scaled as
    \begin{equation}
    \Delta \tau_j = \frac{\Delta \tau_j - \Delta \tau_{\text{mean}}}{\max(\Delta \tau_{\text{max}} - \Delta \tau_{\text{mean}},\Delta \tau_{\text{mean}} - \Delta \tau_{\text{min}})}.
    \end{equation}
    
    All normalized trajectories are subsequently clustered into \(k\) groups using the k-means method, representing \(k\) distinct trajectory modes. The mean and standard deviation of each cluster are then computed and used to parameterize the components of the Gaussian mixture model. The prior distribution \(p_0\) is then defined as
    \begin{equation}
    p_0 : =  \sum_{k=1}^{K} \pi_k \mathcal{N}(\mu_k, \sigma_k^2 \cdot I),
    \end{equation}
    where \(K\) denotes the number of Gaussian components, \(\pi_k\) represents the mixing coefficient for the \(k\)-th component, \(\mu_k\) is the mean, and \(\sigma_k^2\) is the variance for the \(k\)-th Gaussian component. In this work, we fix  \(\pi_k=1\). Exploring how to model more optimal parameters remains an interesting direction for future research.

    \begin{table*}[htbp]
    \setlength{\tabcolsep}{7pt}
    \caption{\textbf{Performance on the NAVSIMv1 navtest benchmark.} ``C'' denotes camera, and ``L'' denotes LiDAR. \textit{*} indicates results reported from the official papers.}
    \vskip -0.1in
    \label{tab:navsimv1}
    \begin{center}
    \begin{small}
    \begin{tabular}{l|ccc|cc|ccc|c}
    \toprule
    Method 
    & Venue 
    & Input 
    & Img. Backbone 
    & NC\(\uparrow\) 
    & DAC\(\uparrow\) 
    & TTC\(\uparrow\) 
    & Comf.\(\uparrow\) 
    & EP\(\uparrow\) 
    & \cellcolor[HTML]{DCDCDC}PDMS\(\uparrow\) \\
    \midrule
    
    TransFuser\cite{transfuser}  
    & IEEE TPAMI 
    & C \& L
    & ResNet-34 
    & 97.7 
    & 92.8 
    & 92.8 
    & \textbf{100} 
    & 79.2 
    & \cellcolor[HTML]{DCDCDC}84.0  \\
    
    VADv2\cite{VADv2} 
    &  arXiv 2024
    & C \& L
    & ResNet-34 
    & 97.2 
    & 89.1 
    & 91.6 
    & \textbf{100} 
    & 76.0 
    & \cellcolor[HTML]{DCDCDC}80.9  \\
    
    Hydra-MDP\cite{Hydra-MDP}  
    &  arXiv 2024
    & C \& L 
    & ResNet-34 
    & 98.3 
    & 96.0 
    & 94.6 
    & \textbf{100} 
    & 78.7 
    & \cellcolor[HTML]{DCDCDC}86.5  \\
    
    GoalFlow*\cite{goalflow} 
    & CVPR 2025
    & C \& L
    & ResNet-34 
    & 98.3 
    & 93.8 
    & 94.3 
    & \textbf{100} 
    & 79.8 
    & \cellcolor[HTML]{DCDCDC}85.7  \\
    
    DiffusionDrive \cite{DiffusionDrive}  
    & CVPR 2025 
    & C \& L 
    & ResNet-34 
    & 98.2 
    & 96.2 
    & 94.7 
    & \textbf{100} 
    & 82.2 
    & \cellcolor[HTML]{DCDCDC}88.1  \\

    WoTE \cite{wote}  
    & ICCV 2025 
    & C \& L 
    & ResNet-34 
    & 98.5 
    & 96.8 
    & 94.9 
    & 99.9 
    & 81.9 
    & \cellcolor[HTML]{DCDCDC}88.3 
    \\

    UniAD\cite{UniAD} 
    & CVPR 2023 
    & C 
    & ResNet-34 
    & 97.8 
    & 91.9 
    & 92.9 
    & \textbf{100} 
    & 78.8 
    & \cellcolor[HTML]{DCDCDC}83.4  \\

    World4Drive\cite{zheng2025world4drive}  
    & ICCV 2025
    & C
    & ResNet-34 
    & 97.4 
    & 94.3 
    & 92.8 
    & \textbf{100} 
    & 79.9 
    & \cellcolor[HTML]{DCDCDC}85.1  \\
    
    Epona \cite{zhang2025epona}  
    & ICCV 2025 
    & C
    & Transformer
    & 97.9 
    & 95.1 
    & 93.8 
    & 99.9 
    & 80.4 
    & \cellcolor[HTML]{DCDCDC}86.2  \\
    
    \midrule
    
    MeanFuser(Ours) 
    & -
    & C
    & ResNet-34 
    & \textbf{98.6} 
    & \textbf{97.0} 
    & \textbf{95.0} 
    & \textbf{100} 
    & \textbf{82.8} 
    & \cellcolor[HTML]{DCDCDC}\textbf{89.0}  \\
    
    \bottomrule
    \end{tabular}
    \end{small}
    \end{center}
    \vskip -0.1in
    \end{table*}

    \begin{table*}[htbp]
    \setlength{\tabcolsep}{8pt}
    \caption{\textbf{Performance of the NAVSIMv2 navtest benchmark.} $^\dagger$ denotes testing with the official checkpoint}
    \vskip -0.1in
    \label{tab:navsimv2}
    \begin{center}
    \begin{small}
    \begin{tabular}{l|ccccccccc|c}
    \toprule
    Method 
    & NC\(\uparrow\) 
    & DAC\(\uparrow\) 
    & DDC\(\uparrow\) 
    & TLC\(\uparrow\) 
    & EP\(\uparrow\)
    & TTC\(\uparrow\) 
    & LK\(\uparrow\) 
    & HC\(\uparrow\) 
    & EC\(\uparrow\) 
    & \cellcolor[HTML]{DCDCDC}EPDMS\(\uparrow\) \\
    \midrule
    
    Ego Status MLP\cite{admlp} 
    & 93.1
    & 77.9
    & 92.7
    & 99.6 
    & 86.0 
    & 91.5 
    & 89.4
    & \textbf{98.3} 
    & 85.4
    & \cellcolor[HTML]{DCDCDC}64.0 \\
    
    TransFuser\cite{transfuser}  
    & 96.9
    & 89.9
    & 97.8
    & 99.7
    & 87.1
    & 95.4
    & 92.7
    & \textbf{98.3}
    & 87.2
    & \cellcolor[HTML]{DCDCDC}76.7\\
    
    Hydra-MDP++\cite{hydra-MDP++}  
    & 97.2
    & \textbf{97.5}
    & 99.4
    & 99.6 
    & 83.1
    & 96.5
    & 94.4
    & 98.2
    & 70.9
    & \cellcolor[HTML]{DCDCDC}81.4 \\
    
    DriveSuprim\cite{drivesuprim}  
    & 97.5
    & 96.5
    & 99.4
    & 99.6
    & \textbf{88.4}
    & 96.6
    & 95.5 
    & \textbf{98.3} 
    & 77.0
    & \cellcolor[HTML]{DCDCDC}83.1 \\
    
    DiffusionDrive $^\dagger $\cite{DiffusionDrive}  
    & 98.2
    & 96.3
    & 99.4
    & \textbf{99.8} 
    & 87.4 
    & \textbf{97.4}
    & 97.0
    & \textbf{98.3} 
    & 87.7
    & \cellcolor[HTML]{DCDCDC}88.3\\
    \midrule
    
    MeanFuser(Ours)  
    & \textbf{98.3}
    & 97.2
    & \textbf{99.6}
    & \textbf{99.8}
    & 87.6
    & \textbf{97.4}
    & \textbf{97.3}
    & \textbf{98.3}
    & \textbf{88.2}
    & \cellcolor[HTML]{DCDCDC}\textbf{89.5}\\
    
    \bottomrule
    \end{tabular}
    \end{small}
    \end{center}
    \vskip -0.2in
    \end{table*}

    \begin{table}[htbp]
        \setlength{\tabcolsep}{2pt}
        \caption{\textbf{Model parameter size, inference speed, and performance.} \textbf{Bold} and \underline{underlined} values denote the best and second-best results, respectively. \textit{``Dim"} indicates the number of hidden neurons in the model, \textit{``FPS"} represents the median inference speed measured on a single NVIDIA H20 GPU over multiple runs, and \textit{``Plan FPS"} refers to the inference speed of trajectory planning excluding the perception encoder.}

        \vskip -0.1in
        \label{tab:ablation_fps}
        \begin{center}
        \begin{small}
        \begin{tabular}{l|cc|ccc}
        \toprule
        Method
        & Dim
        & Params (M)
        & PDMS $\uparrow$
        & Plan FPS $\uparrow$
        & FPS $\uparrow$ \\
        \midrule
        
        TransFuser
        & 256
        & 55.9
        & 84.0
        & \textbf{3934}
        & \textbf{63} \\
        
        GoalFlow
        & 256
        & 62.3
        & 85.7
        & 11
        & 10 \\
        
        Hydra-MDP
        & 256
        & 65.2
        & 86.5
        & 25
        & 20 \\
        
        DiffusionDrive
        & 256
        & 60.7
        & \uline{88.1}
        & 75
        & 39 \\

        \midrule
        MeanFuser
        & 128
        & 54.6
        & \textbf{89.0}
        & \uline{434}
        & \uline{59} \\
        \bottomrule
        \end{tabular}
        \end{small}
        \end{center}
        \vskip -0.3in
    \end{table}
    
    \subsection{Multi-modal Trajectories Sample}

        We train a lightweight network for estimating the average velocity, \( u_{\theta}(\tau_t, r, t \mid \mathbf{c}) \), which maps noise sample \( \tau_0 \sim p_0 \) to the expert trajectory \( \tau_1 \). The noise samples are used as queries, while the specific scene information \( \mathbf{c} \), along with the encoded time intervals \( r \) and \( t \), are concatenated to serve as the key and value for the Transformer decoder layers. These inputs are processed through LayerNorm and a feedforward neural network (FFN) to integrate scene-specific context. This allows the network to focus on the surrounding agents and lane context, ensuring that the generated trajectory is consistent with the current driving scenario. Accordingly, the loss function of the decoder can be formulated as:
        \begin{equation}
            \mathcal{L}_{flow} = \| u_\theta(\tau_t, r, t|\mathbf{c}) - \text{sg}(u_{\text{tgt}}) \|_1.
        \end{equation}

        During training, we select the Gaussian component nearest to the ground truth (using Euclidean distance) and compute its loss only. During model inference, we sample noise points from each Gaussian component and generate planning trajectories in parallel. Each noise point corresponds to a trajectory of a specific driving mode.

    \begin{table*}[htbp]
        \setlength{\tabcolsep}{6pt}
        \caption{\textbf{Ablation study on the impact of each module.} Base denotes the TransFuser\cite{transfuser} baseline. $\mathcal{M}_0$ replaces the decoder MLP in TransFuser with a MeanFlow model and employs a classifier to score the parallel sampled trajectory proposals. $\mathcal{M}_1$ and $\mathcal{M}_2$ further enhance $\mathcal{M}_0$ by introducing GMN and an ARM, respectively.  $\mathcal{M}_3$ employs an averaging strategy across all sampled traiectories. $N_{\text{proposals}}$ denotes the number of sampled trajectory proposals; $P_{L_2>0.2}$ and $P_{L_2>0.5}$ represent the proportions of scenarios where all proposals have a minimum $L_2$ distance greater than 0.2 and 0.5 from the expert trajectory; and $N_{\text{DAC}=0}$ indicates the number of cases where all proposals leave the drivable area.}
        \vskip -0.1in
        \label{tab:ablation_modules}
        \begin{center}
        \begin{small}
        \begin{tabular}{cl|l|cccc}
        \toprule
        Model
        & Components
        & \cellcolor[HTML]{DCDCDC}PDMS\(\uparrow\) 
        & \(N_{proposals}\) 
        &\(P_{L_2>0.2}\)$\downarrow$
        & \(P_{L_2>0.5}\)$\downarrow$ 
        &\cellcolor[HTML]{DCDCDC}\(N_{DAC=0}\)$\downarrow$\\
        \midrule

        -
        & DiffusionDrive
        & \cellcolor[HTML]{DCDCDC}88.1
        & 20
        & 72.7\%
        & 20.0\%
        & \cellcolor[HTML]{DCDCDC}84\\
        \hdashline
        Base
        & TransFuser
        & \cellcolor[HTML]{DCDCDC}84.0
        & -
        & -
        & -
        & \cellcolor[HTML]{DCDCDC}-\\
        
        $\mathcal{M}_0$
        & Base + vanilla MeanFlow
        & \cellcolor[HTML]{DCDCDC}87.3$_{\textcolor{green!50!black}{+3.3}}$
        & 16
        & 92.8\%
        & 40.6\%
        & \cellcolor[HTML]{DCDCDC}143\\
        
        $\mathcal{M}_1$
        & $\mathcal{M}_0$ + Guassian-mixed Noise (\textbf{GMN})
        & \cellcolor[HTML]{DCDCDC}88.2$_{\textcolor{green!50!black}{+0.9}}$
        & 16
        & 69.3\%
        & 18.5\%
        & \cellcolor[HTML]{DCDCDC}58\\
        
        $\mathcal{M}_2$
        & $\mathcal{M}_1$ + Adpative Reconstruction Module (\textbf{ARM})
        & \cellcolor[HTML]{DCDCDC}\textbf{89.0}$_{\textcolor{green!50!black}{+0.8}}$
        & 17
        & \textbf{67.1\%}
        & \textbf{16.9\%}
        & \cellcolor[HTML]{DCDCDC}\textbf{48}\\
        \midrule

        $\mathcal{M}_3$
        & $\mathcal{M}_1$ + Trajectory Proposals Averaging
        & \cellcolor[HTML]{DCDCDC}71.2$_{\textcolor{red}{-17.8}}$
        & 17
        & 69.1\%
        & 18.0\%
        & \cellcolor[HTML]{DCDCDC}57\\
        
        \bottomrule
        \end{tabular}
        \end{small}
        \end{center}
        \vskip -0.1in
    \end{table*}

    \subsection{Adaptive Reconstruction Module}
    
        Model sampling generates multimodal trajectories \(\{\hat{\tau}_k\}_{k=1}^K\), where \(K\) denotes the number of candidates. All candidate trajectories and \(c_{\text{bev}}\) are processed by a cross-attention layer, and the resulting representation is fed into a Projector to yield the final trajectory \(\hat{\tau}\).
        
        During training, we supervise the ARM parameters with expert demonstrations, and compute the loss as follows:
        \begin{equation}
            \mathcal{L}_\tau = \|\tau - \hat{\tau} \|_1,
        \end{equation}
        where \(\tau\) denotes the ground-truth expert trajectory. The model's training does not rely on evaluation metrics from benchmark.

        The total loss function is a weighted sum of the losses from Meanflow decoder, Mapping and ARM, as follows:
        \begin{equation}
            \mathcal{L} = \lambda_1 \mathcal{L}_\tau + \lambda_2 \mathcal{L}_{flow}  + \lambda_3 \mathcal{L}_{map}
        \end{equation}
        The hyperparameters \(\lambda_1\), \(\lambda_2\), \(\lambda_3\) control the relative importance of each component loss.


\section{Experiments}

    \subsection{Dataset and Metrics}
    
        \textbf{NAVSIMv1}: The NAVSIM dataset\cite{navsim} is an end-to-end planning-oriented subset derived from the nuPlan dataset\cite{nuplan}. During the collection process, static scenarios and constant-speed driving scenarios were excluded, leaving behind a large set of challenging scenarios. NAVSIMv1 uses non-reactive simulations along with closed-loop evaluation metrics. The evaluation metric, \textbf{PDM Score (PMDS)}, is a weighted combination of five sub-metrics: no collisions (NC), drivable area compliance (DAC), time-to-collision (TTC), comfort (Comf.), and Ego Progress (EP). The dataset is divided into fixed sets: navtrain and navtest, which are used for training and testing, respectively.

        \noindent \textbf{NAVSIMv2}: NAVSIMv2\cite{navsimv2} offers more comprehensive evaluation metrics and a reactive simulation environment than NAVSIMv1, enabling performance assessment in challenging scenarios. Model performance is reflected by the \textbf{Extended PDM Score (EPDMS)}, which augments the original PDMS with additional submetrics: driving direction compliance (DDC), traffic light compliance (TLC), lane keeping (LK), history comfort (HC), and extended comfort (EC). 

    \subsection{Implementation Detail}
        
        We apply AdamW with weight decay 0.1, using a cosine-annealed schedule that starts at $2\times10^{-4}$ and includes a 3-epoch warm-up. The network uses a hidden dimension of 128 and emits 4-second trajectories at 2 Hz, yielding eight discrete waypoints. During both training and inference, the noise distribution is parameterized as an 8-component Gaussian mixture (\(K=8\)), and we sample eight trajectories in parallel for each scene.

    \begin{figure*}[ht]
    \vskip 0.2in
        \begin{center}
        \centerline{\includegraphics[width=2.2\columnwidth]{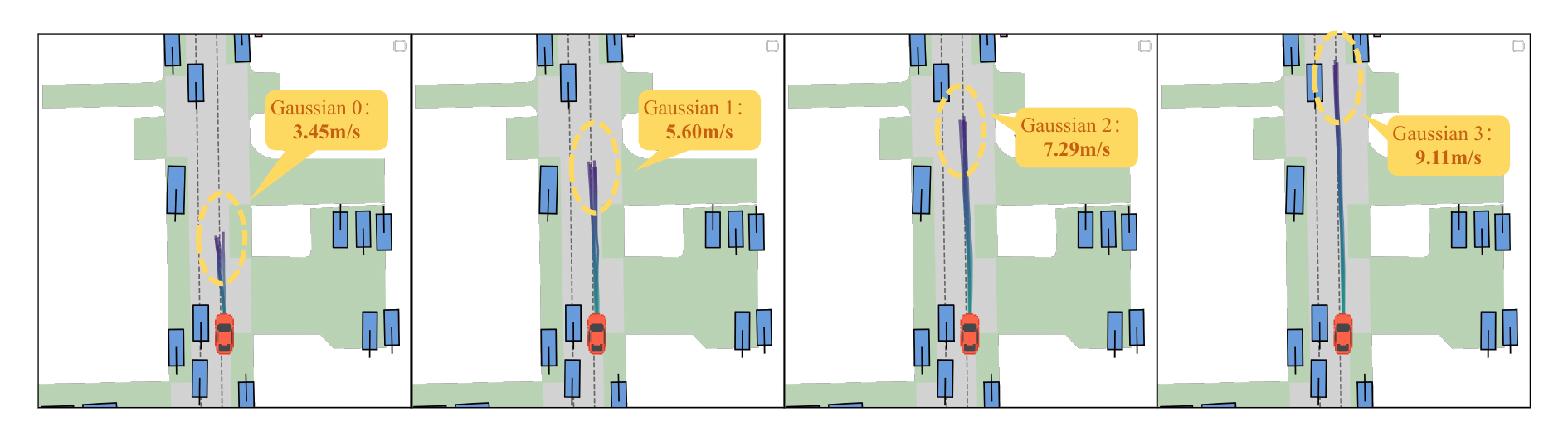}}
        \caption{Visualization of sampling from different Gaussian components. Parallel sampling of trajectories from distinct Gaussian components can generate diverse driving styles, ranging from conservative to aggressive.}
        \label{fig:scenes_style}
        \end{center}
    \vskip -0.2in
    \end{figure*}

    \begin{figure}[ht]
        \begin{center}
        \centerline{\includegraphics[width=1\columnwidth]{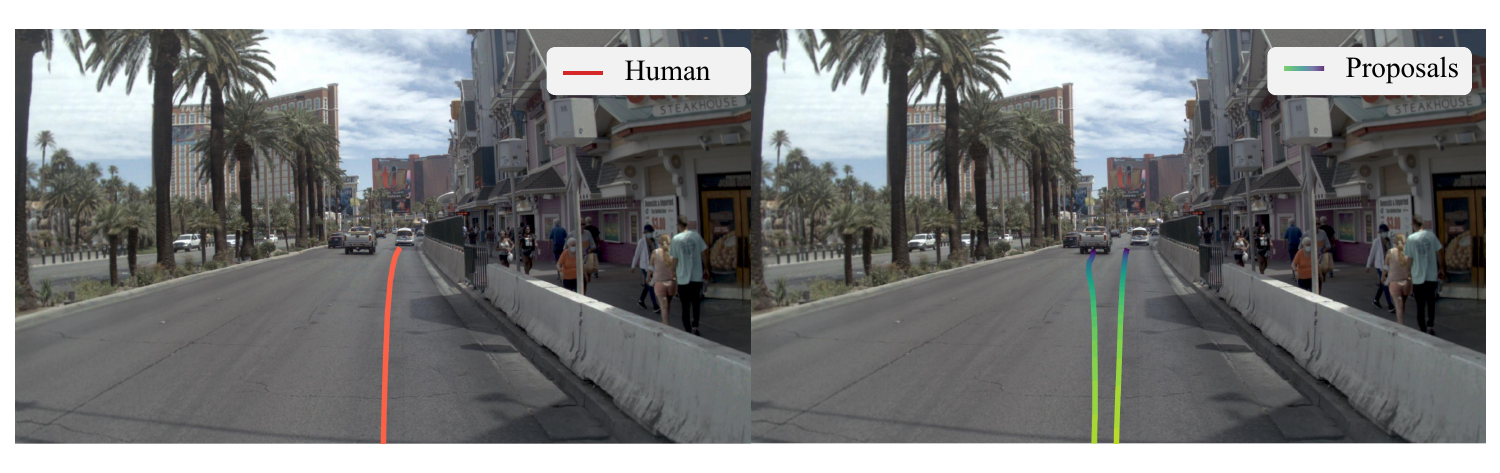}}
        \caption{Visualization of the multimodal trajectory of the model. The left image shows expert demonstration trajectories, while the right image displays the model’s inferred strategies for maintaining a straight path and performing a left lane change, representing two different modes.}
        \label{fig:scenes_multimodal}
        \end{center}
    \vskip -0.4in
    \end{figure}
    
    \subsection{Main Results}

        As shown in Tab.~\ref{tab:navsimv1} and Tab.~\ref{tab:navsimv2}, under a shared ResNet-34 visual backbone, {MeanFuser} achieves the strongest closed-loop performance on both{NAVSIM-v1} and NAVSIM-v2 benchmarks using only RGB visual input (without LiDAR), outperforming all multimodal (C\&L) competitors. On NAVSIMv1, MeanFuser reaches a PDMS of \textbf{89.0}, surpassing the diffusion-based DiffusionDrive by \textbf{+0.9} and the flow based {GoalFlow} by \textbf{+3.3}, while achieving the highest scores across all sub-metrics. On {NAVSIMv2}, which emphasizes generalization and reactive control, further confirms these findings. {MeanFuser} attains state-of-the-art performance with \textbf{EPDMS = 89.5}, showing remarkable gains on the LK and EC metrics and continuous improvement in compliance and safety indicators. Overall, these results demonstrate that explicitly modeling multimodal uncertainty in the trajectory space through the Gaussian mixture formulation, combined with adaptive trajectory reconstruction, enables MeanFuser to achieve superior driving performance under a lightweight vision-only setup while substantially enhancing safety margins and regulatory compliance.
    
        As shown in Tab.~\ref{tab:ablation_fps}, we further compare the performance and inference speed of different models. The results highlight the significant advantage of MeanFuser's lightweight network design. MeanFuser achieves the highest PDMS of 89.0 and an impressive inference speed of 59 FPS, outperforming the Hydra-MDP model on both metrics. The one-step sampling strategy used in MeanFuser eliminates the need for time-consuming iterative sampling. Notably, when using the same perception model, MeanFuser accelerates the trajectory planning module by 39.45× and 5.78× compared to GoalFlow and DiffusionDrive, respectively. This substantially improves inference efficiency, as these baseline methods are more computationally intensive due to their reliance on multiple sampling steps or complex architectures.

    \subsection{Ablation Studies}

    \textbf{The ablation of each core module.} Detailed results are reported in Table \ref{tab:ablation_modules}. The baseline TransFuser model employs an MLP decoder for direct trajectory planning. Based on this, $\mathcal{M}_0$ replaces the MLP with a context-condition MeanFlow trajectory generator, achieving a performance gain of +3.3 PDMS. Subsequently, $\mathcal{M}_1$ further incorporates Gaussian Mixture Noise (GMN) guidance, and $\mathcal{M}_2$ replaces the classifier-based scoring mechanism with the proposed Adaptive Reconstruction Module (ARM), yielding additional improvements of +0.9 PDMS and +0.8 PDMS, respectively. This culminates in a final performance of 89.0 PDMS. In contrast, $\mathcal{M}_3$, which simply averages all trajectory proposals instead of utilizing ARM, suffers a significant performance drop of 17.8 PDMS. This substantial decline indicates that our sampled trajectory proposals capture diverse behaviors rather than collapsing into a single, averaged mode.

    To further evaluate robustness, we analyze the proportion of scenarios where all trajectory proposals deviate from the expert trajectory by more than 0.2 or 0.5 in Euclidean distance, as well as the number of cases where all proposals leave the drivable area ($DAC=0$). Compared to DiffusionDrive, $\mathcal{M}_1$ successfully handles a greater number of challenging scenarios despite utilizing fewer trajectory proposals. This confirms that {GMN provides broader coverage of the trajectory solution space than conventional anchor-based guidance.} Furthermore, a comparison between $\mathcal{M}_1$ and $\mathcal{M}_0$ reveals 85 additional $DAC=0$ cases for the latter, demonstrating that the {GMN sampling method captures a richer diversity of driving strategies compared to standard Gaussian noise sampling}, thereby being able to cover the optimal driving behavior. Finally, by integrating the trajectory regeneration capability of ARM, $\mathcal{M}_2$ reduces the number of $DAC=0$ cases to {48} from 58, underscoring {ARM's effectiveness in generating higher-quality trajectories when all initially sampled proposals are suboptimal.}

    \subsection{Qualitative Analysis}

    \textbf{Multi-modal trajectory porposals.} As shown in Fig.~\ref{fig:scenes_multimodal}, we demonstrate the model’s multi-modal capability in a specific driving scenario, where it samples both a lane-keeping strategy that closely follows the expert demonstration and an alternative strategy that performs a left-lane change.

   \noindent \textbf{Differentiated driving styles.} As shown in Fig.~\ref{fig:scenes_style}, we visualize 8 trajectory proposals in a specific scenario. All 8 trajectories in each image are sampled from the same Gaussian component. As the mean of the Gaussian component increases, the model’s planned trajectory speed rises progressively from 3.45 m/s to 9.11 m/s, reflecting a shift from conservative to aggressive driving behavior. This demonstrates that the Gaussian Mixture Noise sampling approach enables the modeling of differentiated driving styles.

\section{Conclusion}

    This paper presents a novel end-to-end autonomous driving planning framework MeanFuser that overcomes the limitations of discrete vocabulary dependency and achieves an effective balance between one-step sampling and planning performances. We introduce three technical contributions: Gaussian mixture noise, the ``MeanFlow identity", and an adaptive reconstruction module. We further show its effectiveness and provide extensive analysis of the framework's planning performance, computational efficiency, and trajectory diversity.

%% file: sec/X_suppl.tex
\clearpage
\setcounter{page}{1}
\maketitlesupplementary

\section{CARLA Longest6 Benchmark}
\label{sec:rationale}
\textbf{CARLA Longest6 Benchmark:} The CARLA Longest6 Benchmark, introduced by TransFuser\cite{transfuser}, is designed to reduce computational resource consumption and testing time while ensuring a balanced distribution of routes across six towns, with six test routes selected from each town, resulting in a total of 36 routes averaging 1500 meters in length. The benchmark incorporates six distinct weather conditions and six different times of day, and its evaluation metric is the Driving Score (DS), computed as a weighted average of Route Completion (RC) penalized by the Infraction Score (IS).

 \begin{table}[htbp]
        \setlength{\tabcolsep}{5pt}
        \caption{\textbf{Longest6 Benchmark Results.} We show the mean and std for all metrics (RC: Route Completion, IS: Infraction Score. DS: Driving Score).}

        \vskip -0.1in
        \label{tab:carla}
        \begin{center}
        \begin{small}
        \begin{tabular}{c|ccc}
        \toprule
        Method
        & RC $\uparrow$
        & IS $\uparrow$
        & \cellcolor[HTML]{DCDCDC}DS $\uparrow$ \\
        \midrule
        
        Latent TransFuser\cite{transfuser}
        & \textbf{95.18}$_{\pm 0.45}$
        & 0.38$_{\pm 0.05}$
        & \cellcolor[HTML]{DCDCDC}37.31$_{\pm 5.35}$ \\
        
        TransFuser\cite{transfuser}
        & 93.38$_{\pm 1.20}$
        & 0.50$_{\pm 0.06}$
        & \cellcolor[HTML]{DCDCDC}47.30$_{\pm 5.72}$ \\
        
        DiffusionDrive\cite{DiffusionDrive}
        & 94.16$_{\pm 1.46}$
        & 0.69$_{\pm 0.02}$
        & \cellcolor[HTML]{DCDCDC}64.27$_{\pm 2.43}$ \\

        MeanFuser (Ours)
        & 94.65$_{\pm 1.32}$
        & \textbf{0.73}$_{\pm 0.05}$
        & \cellcolor[HTML]{DCDCDC}\textbf{70.08}$_{\pm 3.20}$ \\
        
        \bottomrule
        \end{tabular}
        \end{small}
        \end{center}
    \end{table}

\begin{figure}[ht]
        \begin{center}
        \centerline{\includegraphics[width=1\columnwidth]{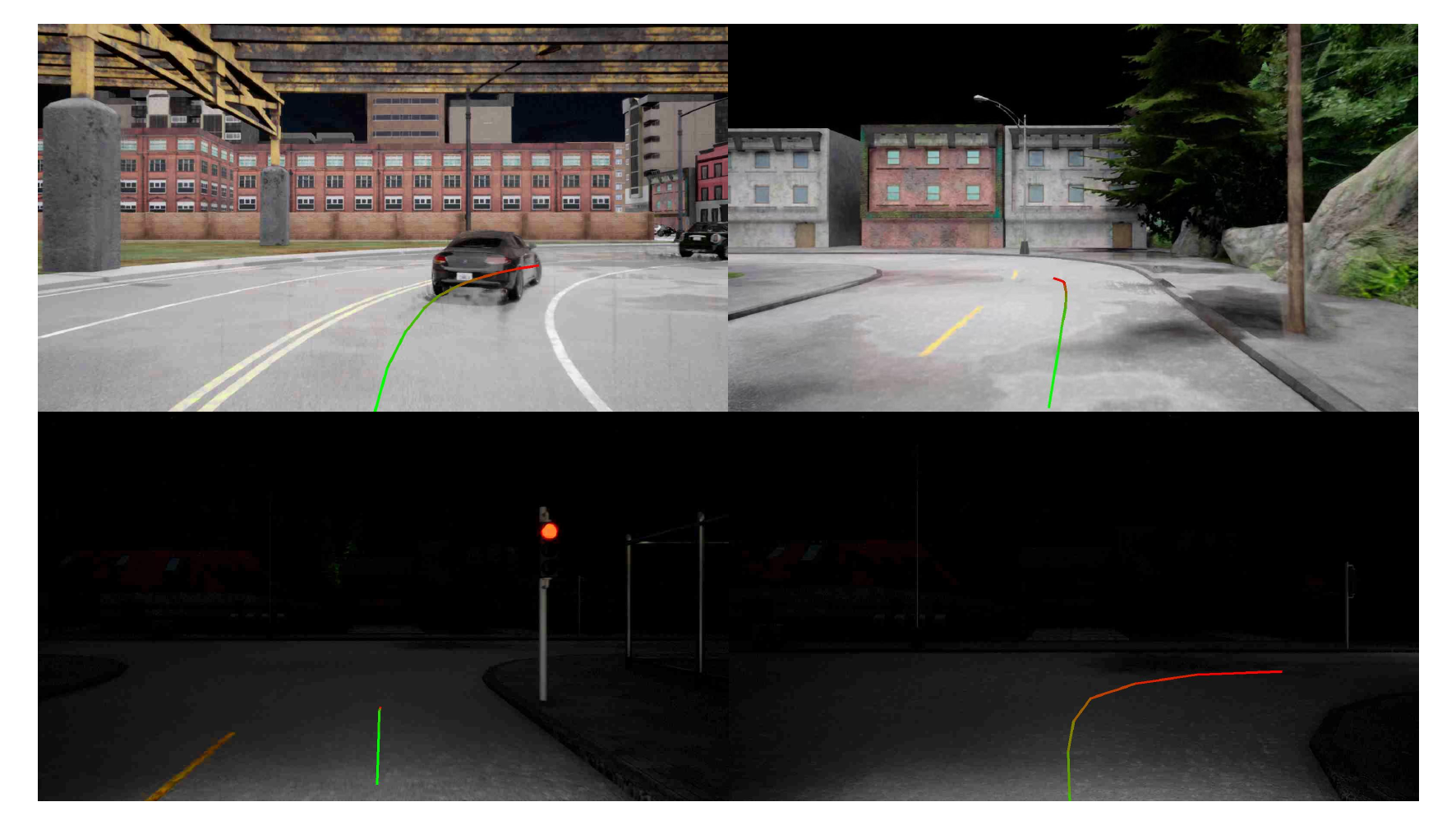}}
        \caption{\textbf{Visualization on the CARLA Longest6 benchmark.} The two images in the upper row and the two in the lower row showcase the planning outcomes, depicting daytime and nighttime conditions, respectively.}
        \label{fig:carla}
        \end{center}
    \end{figure}

    To comprehensively evaluate model performance, we validate our approach on the CARLA Longest6 closed-loop benchmark. As shown in Tab.~\ref{tab:carla}, we conduct three runs for each route to compute the mean and standard deviation. Experimental results demonstrate that the model achieves a Driving Score (DS) that surpasses the unimodal trajectory method TransFuser by 22.78 and outperforms the multimodal method DiffusionDrive by 5.81, confirming its effectiveness and robustness in closed-loop testing. In Fig.~\ref{fig:carla}, we visualize the planning outcomes of our model across diverse scenarios under both daytime and nighttime conditions.

\section{Further Ablation Study}

\begin{figure*}[ht]
\begin{center}
    \centering
    \begin{subfigure}[b]{0.47\textwidth}
        \centering
        \includegraphics[width=\textwidth]{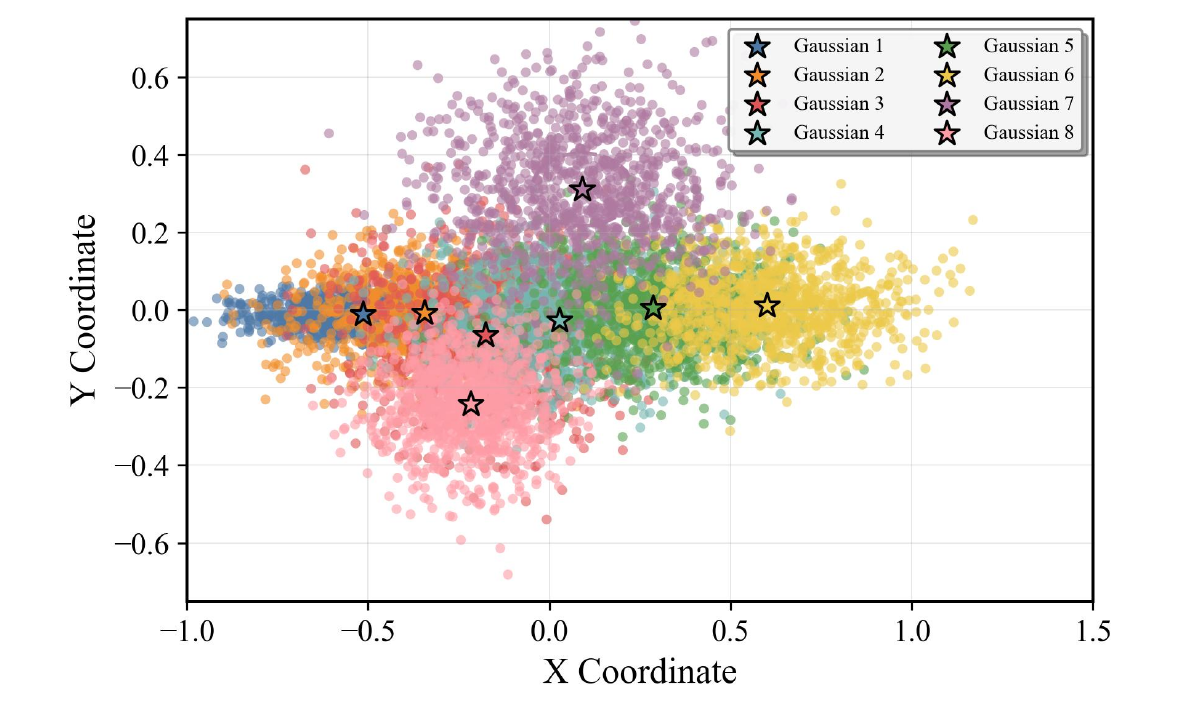}
        \caption{GMN from Data Clustering}
        \label{fig:teaser}
    \end{subfigure}
    \hspace{-0.5cm} 
    \begin{subfigure}[b]{0.47\textwidth}
        \centering
        \includegraphics[width=\textwidth]{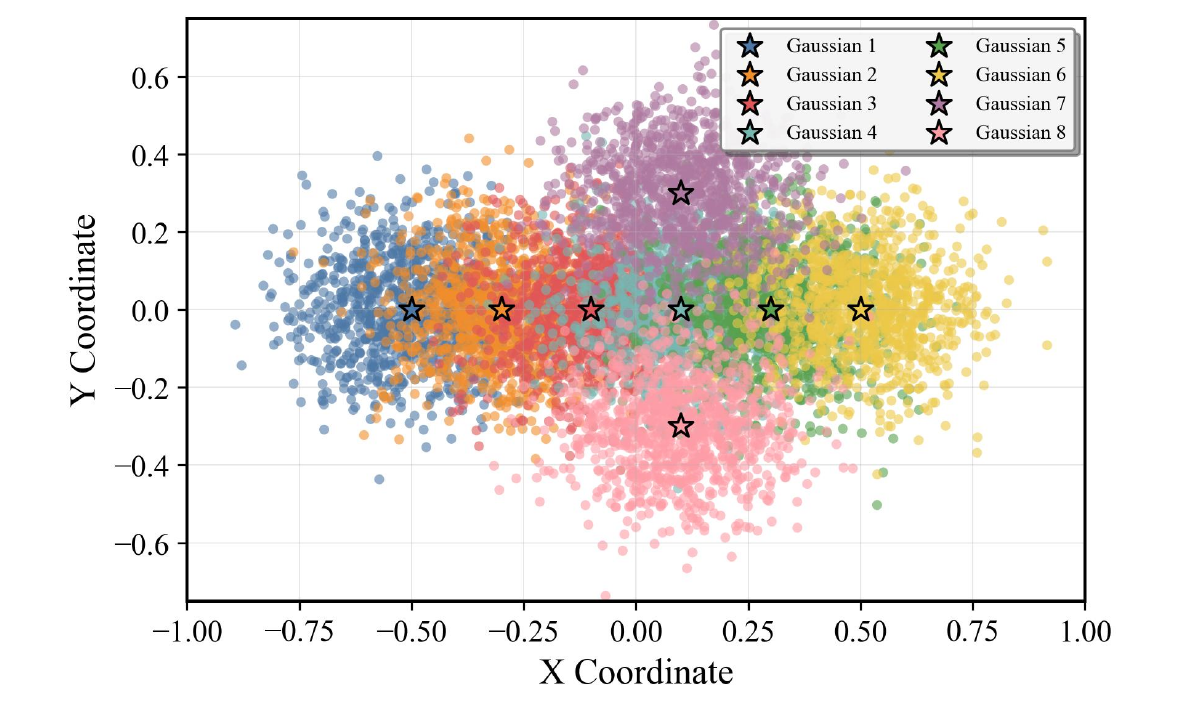}
        \caption{GMN from Manual Design}
        \label{fig:fps}
    \end{subfigure}
    \caption{\textbf{Visualization of alternative approaches for generating Gaussian Mixture Noise (GMN)}. (a) Mean and standard deviation are derived from clustered expert demonstrations in the training set. (b) Mean and standard deviation are obtained through manually design.}
    \vspace{-0.1in}
    \label{fig:anchors}
    \end{center}
    \end{figure*}

    \begin{table}[htbp]
        \setlength{\tabcolsep}{3pt}
        \caption{\textbf{Number of Gaussian components and model performance.} $N_{gaussian}$ denotes the number of Gaussian components in the Gaussian Mixture Noise distribution.}

        \vskip -0.1in
        \label{tab:ablation_gaussian_number}
        \begin{center}
        \begin{small}
        \begin{tabular}{c|ccccc|l}
        \toprule
        $N_{gaussian}$
        & NC $\uparrow$
        & DAC $\uparrow$
        & TTC $\uparrow$
        & Comf. $\uparrow$
        & EP $\uparrow$
        & \cellcolor[HTML]{DCDCDC}PDMS $\uparrow$ \\
        \midrule

        2
        & 98.1
        & 96.9
        & 94.0
        & 100
        & 81.8
        & \cellcolor[HTML]{DCDCDC}88.4$_{\textcolor{red}{-0.5}}$ \\
        
        8
        & 98.6
        & 97.0
        & 95.0
        & 100
        & 82.8
        & \cellcolor[HTML]{DCDCDC}89.0 \\
        
        16
        & 98.1
        & 97.3
        & 93.8
        & 99.9
        & 83.3
        & \cellcolor[HTML]{DCDCDC}88.8$_{\textcolor{red}{-0.2}}$ \\
        
        32
        & 98.0
        & 97.0
        & 94.3
        & 99.9
        & 82.8
        & \cellcolor[HTML]{DCDCDC}88.5$_{\textcolor{red}{-0.5}}$ \\
        \bottomrule
        \end{tabular}
        \end{small}
        \end{center}
    \end{table}

    \textbf{The ablation of the number of Gaussian components.}
    As shown in Tab.~\ref{tab:ablation_gaussian_number}, we present the performance of the model with different numbers of Gaussian components. The model achieves optimal performance with eight Gaussian components. Adding more components beyond this point does not improve results and may even cause a slight decrease in performance. 
    This indicates that eight components provide sufficient capacity to model the trajectory distribution. Further increases lead to each component becoming data-starved, preventing the model from adequately learning the velocity field and resulting in unreliable velocity predictions.

    \begin{table}[htbp]
        \setlength{\tabcolsep}{2.pt}
        \caption{\textbf{Comparison of the GMN Generation Methods.}}

        \vskip -0.1in
        \label{tab:GMN_method}
        \begin{center}
        \begin{small}
        \begin{tabular}{c|ccccc|l}
        \toprule
        Method
        & NC $\uparrow$
        & DAC $\uparrow$
        & TTC $\uparrow$
        & Comf. $\uparrow$
        & EP $\uparrow$
        & \cellcolor[HTML]{DCDCDC}PDMS $\uparrow$ \\
        \midrule

        Data Clustering
        & 98.6
        & 97.0
        & 95.0
        & 100
        & 82.8
        & \cellcolor[HTML]{DCDCDC}89.0 \\

        Manual Design
        & 98.2
        & 97.0
        & 94.1
        & 100
        & 83.0
        & \cellcolor[HTML]{DCDCDC}88.6$_{\textcolor{red}{-0.4}}$ \\
        
        \bottomrule
        \end{tabular}
        \end{small}
        \end{center}
    \end{table}

    \textbf{The ablation study of Gaussian Mixture Noise generation.} To investigate the dependence of model performance on Gaussian Mixture Noise (GMN) generation strategies, we conduct a comprehensive ablation study. As summarized in Tab.~\ref{tab:GMN_method}, we quantitatively compare two GMN construction methods: one derived from clustering expert trajectories in the navtrain dataset (detailed in Section 4.2), and the other manually designed. In our manual design approach, the mean of each Gaussian component is determined by simple heuristic rules while standard deviations are set to a unified fixed value. Experimental results demonstrate that when using manually designed GMN, model performance decreases by only 0.45\% compared to that of the data-driven clustering approach. In contrast, the extreme case where all Gaussian components follow standard normal distributions leads to significant performance degradation. This confirms that our model's effectiveness does not rely on specific fixed datasets. Visual comparisons of the GMN generated by both methods are presented in Fig.~\ref{fig:anchors}.

    \begin{table}[htbp]
        \setlength{\tabcolsep}{5.pt}
        \caption{\textbf{Comparison of multimodality and performance.} (GMN: Gaussian Mixture Noise. $K$: number of multimodal trajectories; $\mathcal{D}$: multimodality metric.)}

        \vskip -0.1in
        \label{tab:multi-modal}
        \begin{center}
        \begin{small}
        \begin{tabular}{c|cccc|l}
        \toprule
        Method
        & GMN
        & $K$
        & PDMS $\uparrow$
        & $\mathcal{D}$ $\uparrow$
        & \cellcolor[HTML]{DCDCDC}$\mathcal{M}_{DP}$ $\uparrow$
        \\
        \midrule

        TransFuser
        & -
        & 8
        & 94.0
        & 0.0
        & \cellcolor[HTML]{DCDCDC}-
        \\

        MeanFuser 
        & $\times $
        & 8
        & 88.3
        & 0.25
        & \cellcolor[HTML]{DCDCDC}22.07
        \\

        MeanFuser
        & $\surd $
        & 8
        & \textbf{89.0}
        & \textbf{0.30}
        & \cellcolor[HTML]{DCDCDC}\textbf{26.70}$_{\textcolor{green!50!black}{+20.84\%}}$
        \\
        
        \bottomrule
        \end{tabular}
        \end{small}
        \end{center}
    \end{table}

    \textbf{The ablation study of Multi-modal planning performence.}

    \begin{figure}[ht]
        \begin{center}
        \centerline{\includegraphics[width=1\columnwidth]{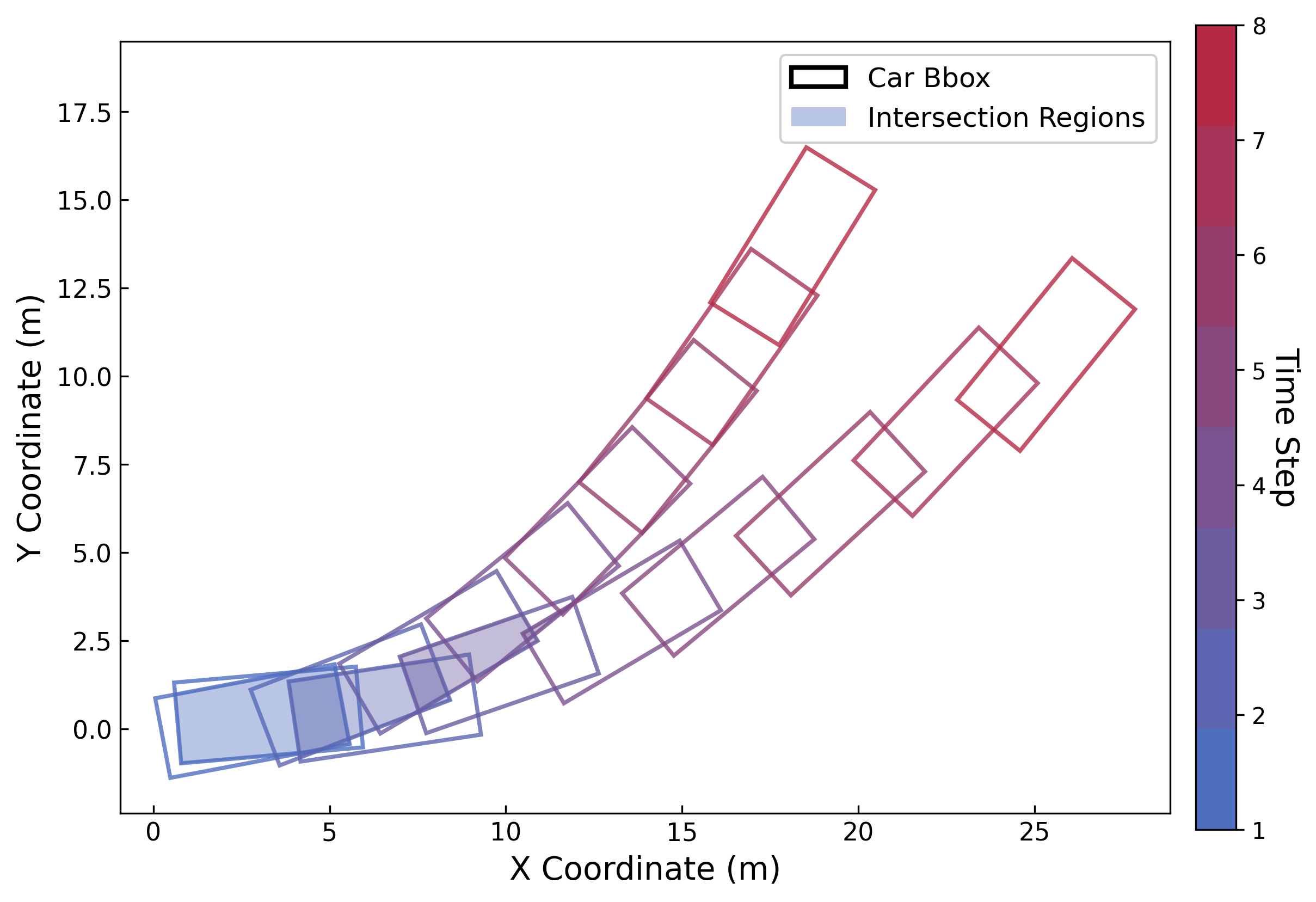}}
        \caption{Visualization of the intersection and union dynamics of bounding boxes (Car Bbox) for ego trajectories across different timesteps.}
        \label{fig:iou}
        \end{center}
    \vskip -0.4in
    \end{figure}

    We employ a mean Intersection-over-Union (mIoU)-based metric $\mathcal{D}$ to quantify the multimodality of planning outcomes. For a set of $K$ trajectories $\{ \tau_k \}_{k=1}^K$, the metric is defined as:
    \begin{equation}
    \mathcal{D} = 1 - \frac{1}{T_f} \sum_{i=1}^{T_f} \frac{ \bigcap_{k=1}^{K} \mathrm{Area}(\hat{\tau}_{ki}) }{ \bigcup_{k=1}^{K} \mathrm{Area}(\hat{\tau}_{ki}) },
    \end{equation}
    where $T_f$ denotes the prediction horizon, $\hat{\tau}_{ki}$ represents the bounding box of the $k$-th trajectory at timestep $i$, and the operators $\bigcap$ and $\bigcup$ calculate the intersection and union of areas across all $K$ trajectories at each timestep $i$, respectively. A higher value of $\mathcal{D}$ indicates greater diversity among the predicted trajectories. For an intuitive understanding, Fig.~\ref{fig:iou} visualizes the bounding boxes at different timesteps along with their intersections.

    To ensure that the observed diversity does not stem from model error, such as trajectory divergence, we introduce the composite metric: 
    \begin{equation}
    \mathbf{\mathcal{M}_{DP}}=\mathcal{D}\times PDMS,
    \end{equation} which provides a unified measure that considers both planning performance and trajectory diversity.

    The ablation studies in Tab.~\ref{tab:multi-modal} demonstrate that the GMN not only enhances the primary performance metric (PDMS) but also significantly increases the diversity ($\mathcal{D}$) of the generated trajectory proposals, resulting in a 20.84\% improvement in the comprehensive evaluation metric $\mathcal{M}_{DP}$.